\documentclass[10pt,twocolumn,letterpaper]{article}

\usepackage{iccv}
\usepackage{times}
\usepackage{epsfig}
\usepackage{graphicx}
\usepackage{amsmath}
\usepackage{amssymb}
\usepackage[accsupp]{axessibility}

\graphicspath{ {./figs/} }               
\usepackage[format=hang]{caption}       
\usepackage[ruled,vlined]{algorithm2e}

\usepackage[pagebackref=true,breaklinks=true,letterpaper=true,colorlinks,bookmarks=false]{hyperref}

\iccvfinalcopy 

\ificcvfinal\fi

\begin{document}

\title{MEAL: Manifold Embedding-based Active Learning}

\author{Deepthi Sreenivasaiah, Johannes Otterbach, Thomas Wollmann\\
Merantix Labs GmbH, Berlin, Germany\\
{\tt\small \{deepthi.sreenivasaiah, johannes.otterbach, thomas.wollmann\}@merantix.com}
}

\maketitle
\ificcvfinal\thispagestyle{empty}\fi

\begin{abstract}
Image segmentation is a common and challenging task in autonomous driving. Availability of sufficient pixel-level annotations for the training data is a hurdle. Active learning helps learning from small amounts of data by suggesting the most promising samples for labeling. In this work, we propose a new pool-based method for active learning, which proposes promising patches extracted from full image, in each acquisition step. The problem is framed in an exploration-exploitation framework by combining an embedding based on Uniform Manifold Approximation to model representativeness with entropy as uncertainty measure to model informativeness. We applied our proposed method to the autonomous driving datasets CamVid and Cityscapes and performed a quantitative comparison with state-of-the-art baselines. We find that our active learning method achieves better performance compared to previous methods.
\end{abstract}

\section{Introduction}
Success of computer vision methods based on machine learning depends heavily on the quality of labeled data available. Image Segmentation is one of the key sub-tasks in computer vision which has made tremendous progress over the past years \cite{recent_changes_segmentation}. Despite this progress, image segmentation methods based on Convolution Neural Networks~(CNNs) \cite{726791} still pose the big hurdle of availability of sufficient pixel-level annotations for the training data \cite{crowdsourcing}. In domains like autonomous driving \cite{9006846} or biomedical image analysis \cite{suggestive_annot, cell_tracking_challenge}, large quantities of unlabeled data is available but the effort required to manually annotate the data pixel-wise is a major bottleneck \cite{crowdsourcing} in terms of cost and time. In such scenarios, active learning~(AL) is a framework where the machine learning model learns from small amounts of data and selects what data it wants to learn from \cite{settles2009active}. In an active learning setup, a model is initially trained on a small labeled dataset. An acquisition function uses the trained model to select new samples from a pool of unlabeled data points, which are then labeled by an expert (oracle). The newly labeled samples are added to the training dataset and a new model is trained using the updated training dataset. This process is repeated for multiple rounds with the training dataset size growing over the time to reach the required performance. Active learning has proven to drastically reduce the amount of labeled data needed to supervise a task like semantic segmentation \cite{cereals,self-consistency}.

\begin{figure}
    \centering
    \includegraphics[width=\linewidth]{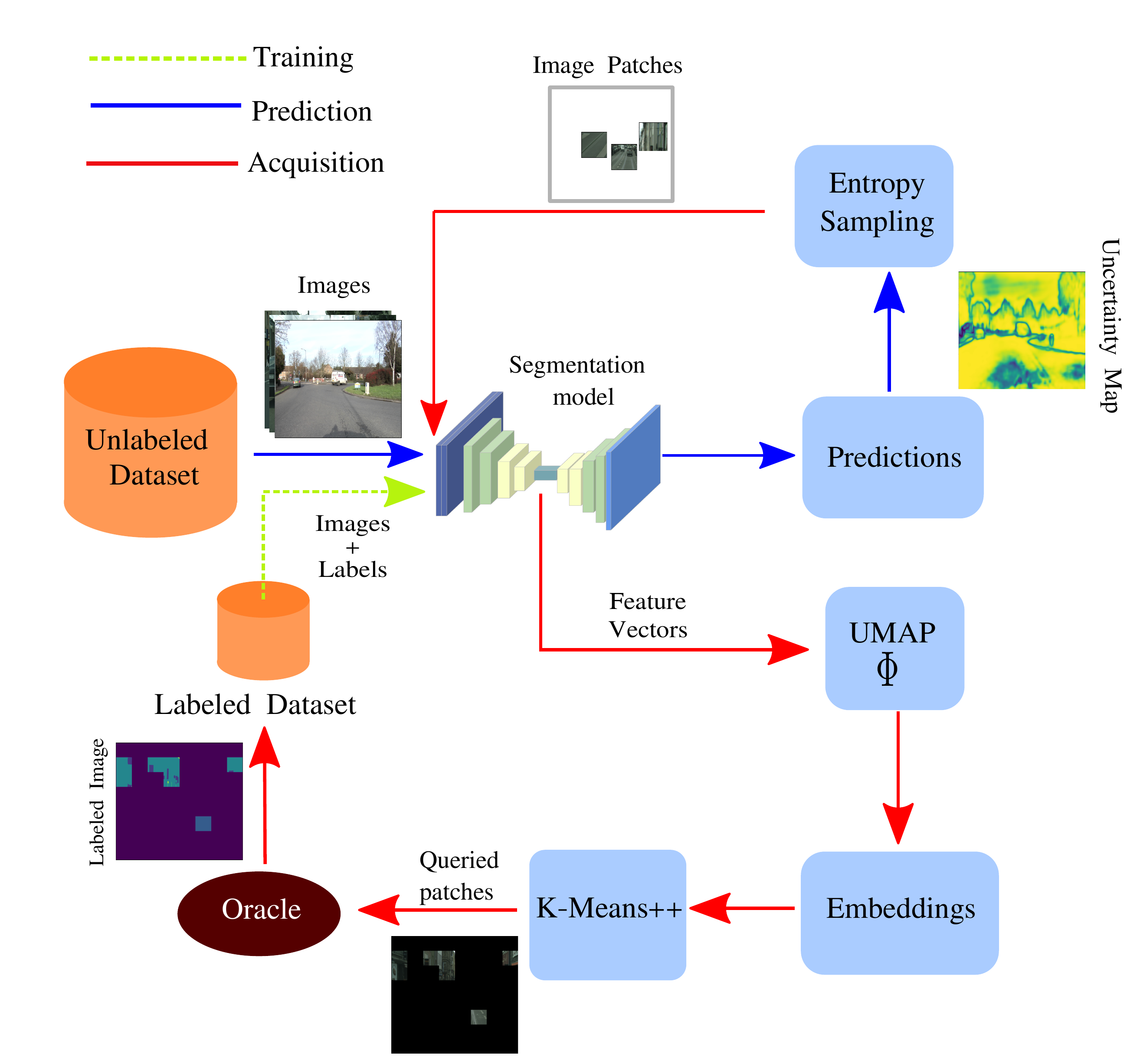}
    \caption{Overview of our proposed method for active learning: MEAL (Color best viewed online).} \label{fig:overview}
\end{figure}

Although annotations for semantic segmentation is a costly and time consuming process compared to other visual recognition tasks, there are very few methods in active learning related to semantic segmentation especially in the field of autonomous driving \cite{cereals,self-consistency}. Currently, many AL methods operating on CNNs focus mostly on image classification tasks.

Most of the AL strategies in the literature involve evaluating either the informativeness \cite{bald,self-consistency}, which focuses on reducing the generalization error or the representativeness \cite{dual} that selects samples to maximize diversity of the unlabeled data. Current AL methods \cite{badge,asknlearn,suggestive_annot} that combine both of these measures do not consider the overall input dataset while designing the representativeness part. Having a trade-off between the two measures is an important consideration in active learning to select high quality and diverse samples. This trade-off could be achieved by using an exploration-exploitation framework \cite{explor-exploi}. In this framework, large regions of unlabeled data space can be explored to select diverse samples and at the same time, exploit smaller regions of unlabeled data space to select highly informative samples. In addition to combining the two measures, one can leverage the spatial coherence of an image to formulate a region-based approach to effectively deal with the large number of samples in a segmentation dataset and the costly annotation of samples for semantic segmentation.

In this work, we present a novel region-based active learning strategy, termed Manifold Embedding-based Active Learning, or MEAL. It selects data samples using informativeness and representativeness measures combined in an exploration-exploitation framework. The overview of our active learning algorithm is represented in Figure~\ref{fig:overview}. For the informativeness measure, we use the Softmax probabilities from the output layer of the model to compute the entropy of image patches. To capture the diversity, a low-level manifold mapping is learned using the feature vectors of all the input images. This manifold embedding is then used to obtain a representative sample of embedding vectors from the selected informative samples. In contrast to previous approaches, our method frames the AL problem as an exploration-exploitation task for region-based active learning in semantic segmentation and can be used in conjunction with other uncertainty methods \cite{self-consistency,bald,ensembles_beluch}. The method learns a low-level manifold mapping using Uniform Manifold Approximations~(UMAP) \cite{umap} to discover a non-linear latent space projection of the data in order to retrieve diverse samples using K-Means++ \cite{K-Means++} in the low dimensional space. To obtain the low-dimensional manifold mapping, we use a MobileNetV2 \cite{MobileNet} backbone pre-trained on ImageNet \cite{ILSVRC15} that is part of our full DeepLabV3 \cite{deeplabv3} segmentation model. We compare our approach to several uncertainty-based active learning methods on the CamVid \cite{camvid} and Cityscapes \cite{Cordts2016Cityscapes} datasets for autonomous driving.

In summary our contributions are:
\begin{itemize}
    \item We rephrase the informativeness-representativeness trade-off of Active Learning in terms of the exploration-exploitation framework
    \item We introduce UMAP-based clustering methods to improve representativeness sampling
    \item We combine the new clustering with existing entropy-based informativeness sampling
    \item We present experimental evidence for the effectiveness of the combined MEAL method based on evaluations on CamVid and Cityscapes datasets.
\end{itemize}

The remainder of this paper is structured as follows. In Section~\ref{sec:related_work}, we review existing AL approaches for image classification and image segmentation. Our AL strategy’s theoretical foundations are outlined and formalized in Section~\ref{sec:meal}. In Section~\ref{sec:experimental}, we present experiments with our AL strategy followed by discussion and future work in Section~\ref{sec:conclusion}.

\section{Related Work} \label{sec:related_work}

Active Learning is a well explored topic in machine learning problems where labeled data is scarce or difficult to obtain \cite{settles2009active,2face_AL}. Settles \cite{settles2009active} provides a rich reference collection of commonly used query strategy frameworks. He outlines several different settings for active learning, among those \textit{pool-based} approaches \cite{settles2009active} have gained more traction in recent times which assumes that there is a small set of labeled data and a large pool of unlabeled data available.

In terms of sample selection, active learning strategies can broadly be classified into uncertainty- \cite{uncertaintysampling} and diversity-based methods. In the former approach, the query strategy is based on some kind of informativeness measure aiming to reduce the model's uncertainty. The intuition is that if a model is highly uncertain about its predictions on an unlabeled sample, then accommodating such a sample into the training is beneficial for the model. Culotta \& McCallum \cite{least_confidence} propose a method which selects unlabeled samples on which the classifier is least confident. Gal et al. \cite{bald} take a different approach by adapting Bayesian deep learning into the active learning framework. Their technique,  Monte Carlo Dropout, performs stochastic forward passes to approximate the posterior distribution which induces the uncertainty in weights giving rise to prediction uncertainty on the unlabeled samples. However, Beluch et al. \cite{ensembles_beluch} prove that their ensemble method perform better in obtaining more calibrated predictive uncertainties which forms the basis for uncertainty-based AL. Diversity-based approaches query a subset of unlabeled data that represents the whole unlabeled dataset covering the underlying data distribution. Sener et al. \cite{coreset} address the problem of AL as a coreset construction by minimizing the $\ell_2$ distance between the activations of fully connected layers of queried point and unlabeled dataset. Another popular approach to query representative samples is running a clustering algorithm like K-Means \cite{kmeans} on the unlabeled dataset and selecting the samples equally from the clusters. This concept was demonstrated in \cite{clustering_al} using spectral clustering.
In \cite{dual}, the authors present an unsupervised active learning approach by selecting representative samples. The method focuses on learning a non-linear embedding to transform the input into a latent space where samples are selected using K-Means clustering \cite{kmeans}.

The uncertainty-based methods \textit{exploits} the input data space to select samples close to decision boundary while completely ignoring the representativeness aspect i.e., it fails to \textit{explore} the larger regions of input data space to select diversified samples. Hence using such a method by itself does not improve performance \cite{explor-exploi}.

On the other hand, there are hybrid methods which chose to combine uncertainty and diversity-based methods \cite{badge,asknlearn,suggestive_annot}. A query strategy called Batch Active learning by Diverse Gradient Embeddings~(BADGE) \cite{badge} combines both diversity and uncertainty methods for batch acquisition by measuring uncertainty through gradient embeddings and diversity through K-Means++ initialization algorithm. Further improvement was made to this method in \cite{asknlearn}, where the authors proposed to calibrate the uncertainties by incorporating a calibration metric into the training objective and add a data augmentation technique to remove confirmation bias during pseudo labelling, while retaining the core active learning querying strategy as in \cite{badge}. In semantic segmentation, this kind of hybrid method can be seen in \cite{suggestive_annot} where the uncertainty is computed using bootstrapping and representative samples are selected using cosine similarity between the unlabeled data samples and informative samples.

\textbf{Uniform Manifold Approximation and Projection (UMAP):} UMAP \cite{umap} is a graph-based non-linear dimensionality reduction method which builds a k-neighbor graph in the original data space and finds a similar graph in the lower dimension. The underlying assumption is the existence of a manifold on which the data is uniformly distributed and preserving the structure of this manifold is a primary goal. We use UMAP to learn the lower dimensional representation of the feature vectors. The main reason is that the graph constructed under this assumption preserves both the local and global structures.

\textbf{Active Learning in Semantic Segmentation:} Compared to image classification, the problem of active learning is less explored in the field of semantic segmentation. Mackowiak et al. \cite{cereals} make use of region-based query strategy where information map is computed as the model's prediction, to extract high informative regions from the image. They train a cost model which predicts the number of clicks needed to annotate each image. Another recent work in semantic segmentation is \cite{self-consistency} which selects image regions that are highly uncertain under equivariant transformation. Siddiqui et al. \cite{viewal} used viewpoint entropy of multiple views of each image as an uncertainty measure based on inconsistencies in the prediction across different views. Casanova et al. \cite{reinforcedAL} proposed a query strategy based on deep reinforcement learning, where an agent learns a policy to select a subset of image regions to label. These existing methods that are related to semantic segmentation make use of only uncertainty based process to query the data and retrain the model.

\section{Manifold Embedding-based Active Learning} \label{sec:meal}

\begin{figure}[hb]
    \centering
    \includegraphics[width=\linewidth]{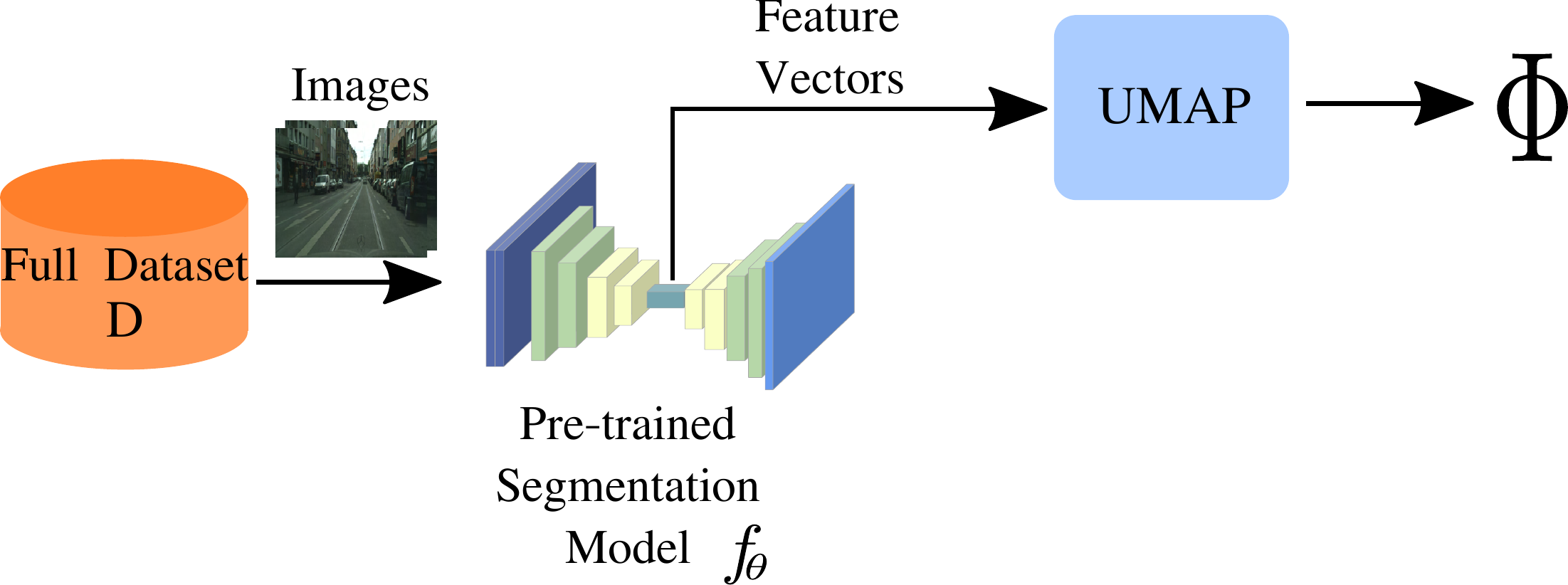}
    \caption{A low dimensional manifold representation $\Phi$ is learned with an UMAP trained on the overall unlabeled dataset $D$ using a segmentation model $f_{\theta}$ loaded with pre-trained ImageNet weights (Color best viewed online).}
    \label{fig:manifold}
\end{figure}

Considering the success of querying the regions of image rather than the entire image in the active learning methods that are tailored for semantic segmentation \cite{self-consistency, cereals, reinforcedAL}, our method is also formulated to select image regions, based on a pre-determined tiling grid, during each acquisition step. 

In our work, we present a novel active learning method that utilizes the exploration-exploitation framework in our query strategy, to achieve high performance on a segmentation model with a low number of labeled samples. The core idea of our approach \textit{Manifold Embedding-based Active Learning (MEAL)} is to utilize the informativeness and representativeness in an exploration-exploitation framework so that the selected samples not only help in reducing the model's uncertainty but are also highly representative for the underlying data distribution. MEAL starts by training an UMAP \cite{umap} across all input data to learn the manifold representation. This manifold representation is then used to construct a low-dimensional embedding of unlabeled samples to enable easier clustering. Converting the entire data space into this low dimensional embedding is computationally intense. To reduce the computational demands we split the sampling process into two steps. In the first step (i) we select a larger subset of highly informative samples using entropy-based uncertainty sampling and obtain the low-dimensional embeddings of these samples using the trained UMAP. In the second step (ii) we use the K-Means cluster centroid initialization algorithm \cite{K-Means++} to compute the cluster centroids which we then add as new samples to the training dataset after labeling. In conjunction these two steps fulfill both the informativeness and the representativeness objective. A challenge with this approach is to select the right number of cluster centroids. 

In the following, we describe the main components of our algorithm, uncertainty sampling and representative sampling, along with the notations in more detail.

\textbf{Notations.} Given the input data $x \subset X \in \mathbb{R}^{h \times w}$ where $h, w$ are the spatial dimensions of the image, labels $c  \in \{1, 2, ..., C\}$ with $C$ classes and label map $y \subset Y \in \mathbb{R}^{h \times w}$, 
let $D_{L}=\left\{x, y\right\}_{i=1} ^{N_{L}}$, $D_{U}=\left\{x\right\}_{i=1} ^{N_{U}}$ be the small labeled and a large pool of unlabeled data, respectively. $D \in \left\{D_U \cup D_L\right\}$ together forms the entire dataset. Let $f_{\theta}: X \rightarrow Y$ be the segmentation model with learnable parameters $\theta$, and the manifold representation learned using UMAP be denoted by $\Phi$ where $\Phi: \mathbb{R}^{d} \rightarrow \mathbb{R}^{d'}$, $d > d'$. The task of our active learning query strategy is to select $N$ image patches at each acquisition step to be labeled by an oracle. This process is iterated for $K$ steps and the selected samples are added to $D_{L}$ dataset along with their corresponding labels. 

\subsection{Exploration-Exploitation Framework}
When querying samples for labeling, there is often a trade-off to select samples which reduce model uncertainty or a subset of samples which represents the overall dataset. This is analogous to exploration and exploitation of the input data space. Exploitation, in the context of active learning corresponds to selecting uncertain samples from an already sampled data space while exploration refers to selecting samples in non-sampled areas of input space. Using UMAP \cite{umap} in our active learning query procedure, we learn a low dimensional manifold of the image patches from the entire input data space which preserves the local neighbourhood structure of the image patches, thus enabling to perform exploration in the lower dimensional space.

\subsection{Learning the Dataset Manifold Approximation}

The curse of dimensionality is an inherent problem when clustering high-dimensional data\cite{bishop_prml}. To address this issue and at the same time preserve the global context of the dataset, we employ UMAP \cite{umap} as a dimensionality reduction technique before clustering the data. Intuitively, this has two effects: (i) we increase the speed of the clustering algorithm by a factor proportional to the reduction in dimensionality and (ii) we increase the robustness of the clusters, since we are eliminating spurious correlations through projecting them into lower dimensionality. It should be pointed out that UMAP preserves local and global structure of the dataset topology and the resulting clusters in low dimensions are hence expected to represent relevant high dimensional clusters. For sake of simplicity, we limit the dimension of the embeddings to two in all our experiments. To train the UMAP embedding map, we make use of the entire dataset using the features obtained from feeding the pre-computed image patches through the segmentation model backbone. In a pool-based setting, we have access to the entire unlabeled dataset at the beginning of active learning process and hence the patch-dataset includes labeled and unlabeled samples and thus information about the whole dataset is accounted for inside UMAP.

\subsection{Uncertainty Sampling}

For uncertainty sampling, we compute the entropy as an uncertainty measure \cite{uncertaintysampling}. Once the segmentation model is trained using the initial $D_{L}$ labeled set, MEAL computes the prediction of all images from the unlabeled data pool $D_{U}$ using the trained model $f_{\theta}$.  $P_{c}^{(u, v)}$ is the Softmax probability of the pixel $(u, v)$ belonging to class $c$ for an image $I_{i}$. The uncertainty map $H_{i}^{(u, v)}$ at each pixel position $(u, v)$ of $i^{th}$ image $I_{i}$ is computed by aggregating probabilities of all the classes as follows.
\begin{align}
    H_{i}^{(u, v)}=- \sum_{c=1}^{C} P_{c}^{(u, v)} \log P_{c}^{(u, v)}
\end{align}
The uncertainty map will have high entropies~(uncertainty) in those regions where the current model is not certain on the predictions and this sampling step selects such image regions. 

Uncertainty maps have a high degree of spatial coherence, i.e. neighboring pixels are likely to have similar (small) entropy as they likely are in the same predicted class where the model is certain. To reduce labeling costs (as well as model compute intensity), we hence select image regions or patches with high entropy to be labelled. For this we divide the uncertainty map $H_{i}$ into rectangular patches of fixed dimension and aggregate the entropy of all pixels in a patch to obtain one uncertainty score $H^{m}_{i}$ for each patch $m$. We then select the region among all the unlabeled dataset that has the highest uncertainty score. This step is repeated for $N_{I}$ times to obtain $N_{I}$ informative samples. 
The selected regions are removed from the unlabeled dataset and processed further in the second step of the query strategy.

\subsection{Representative Sampling}

In the second step of our active learning query strategy, we select a representative subset from the pre-selected informative samples. To do so we obtain feature vectors from the pre-compiled image patches. More formally, let $I_{i,m}$ be the $m^{th}$ region of $i^{th}$ image, then the corresponding feature vector 
\begin{align}
    F_{i,m} = f_{\theta}(I_{i,m})    
\end{align}
is obtained using the trained backbone of the segmentation model $f_{\theta}$. We then feed these features through the pre-computed UMAP $\Phi$ to transform them into the low-dimensional embedding space. We denote these low-dimensional embeddings by 
\begin{align}
    G_{i,m} = \Phi(F_{i, m}).    
\end{align}
Finally, we use the K-Means cluster centroid initialization algorithm (K-Means++ \cite{K-Means++}) to obtain cluster centroids that are used to augment the training dataset. The detailed pseudocode of MEAL is outlined in Algorithm~\ref{algo:meal}.

\begin{algorithm}
	\DontPrintSemicolon
	\textbf{Initialize}: Unlabeled set $D_{U}$, Acquisition steps $K$, Number of query patches $N$, Number of Informative patches $N_{I}$, Initial train set $D_{L}$, Unlabeled set $D_{U}$, Query set $D_{Q}=\varnothing$\\
    \textbf{Step 0}: Train a UMAP $\Phi$ using feature vectors $F$ computed from a segmentation model $f_{\theta}$ with pre-trained ImageNet weights on ($D_{L}, D_{U}$) \\
	\For{$k \gets 1$ \KwTo $K$}{
	    Train a model $f_{\theta}$ on $D_{L}$\\
	    \textbf{Query Procedure:} \\
	    \textbf{Step 1} : Compute pixel-wise Entropy $H_{i}^{(u, v)}$ of image $I_{i}$ in $D_{U}$ using trained $f_{\theta}$ 
		\qquad $H_{i}^{(u, v)}=- \sum_{c=1}^{C} P_{c}^{(u, v)} \log P_{c}^{(u, v)}$  \\
		Select $N_{I}$ image patches with highest entropy $H_{i}$\\
		$I = \varnothing$; $m=1$\\
        \Repeat{$m = N_{I}$; $m\gets m+1$}{
        $I_{i,m} \gets \max{(H)}$\;
        $I \gets I \cup I_{i,m}$\;
        $H \gets H \setminus I_{i,m}$ 
        }
		\textbf{Step 2} : Compute image patch feature vectors $F_{i, m}$ of all $I_{i,m}$ in $I$ starting with $G = \varnothing$ as\\
        \qquad $F_{i,m} = f_{\theta}(I_{i,m})$ \\
		Compute embeddings of the feature vector as \\
		\qquad $G_{i,m} \gets \Phi(F_{i,m})$ \\
        \qquad $G \gets G \cup G_{i,m}$\;
		Run K-Means++ on all embeddings $G$ and select $N$ cluster centroids to form the query set $D_{Q}$ \\
		\textbf{Update:} $D_{U} \gets D_{U} - D_{Q}$, $D_{L} \gets D_{L} \cup D_{Q}$
	 }
	\Return Final training set $D_{L}$
	\caption{MEAL: Manifold Embedding-based Active Learning pseudocode}
	\label{algo:meal}
\end{algorithm}

\subsection{MEAL with Fit Transform}
To interpolate between the full MEAL method and the Entropy baseline, we introduce MEAL with Fit Transform (MEAL-FT). In this variant we do not train a UMAP \cite{umap} using the entire dataset to learn a transformation $\Phi$. Instead the transformation is learned on the informative samples only during each acquisition step and then it is used to transform those samples into low-dimensional embeddings. This modification intends to evaluate if the additional step of incorporating the global context is required. While the original method MEAL carries information of entire dataset in the form of $\Phi$, MEAL-FT does not have this additional information.

\section{Experiments} \label{sec:experimental}
\begin{figure*}[t]
\includegraphics[width=\linewidth]{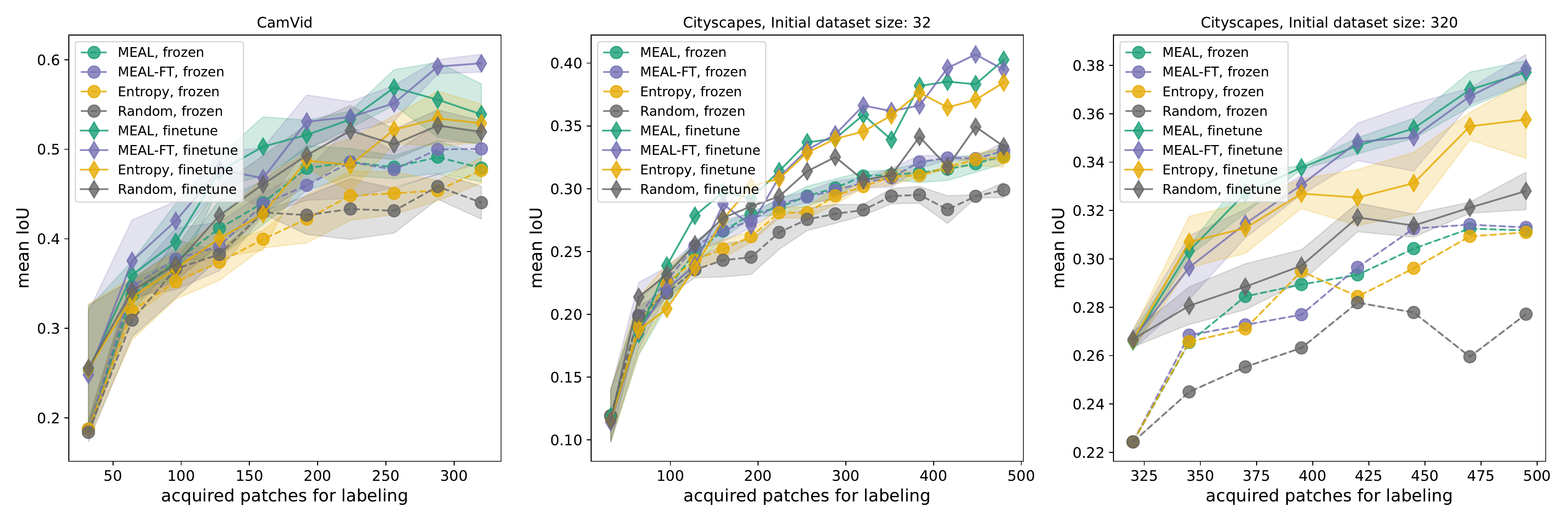}
\caption{Performance of MEAL in comparison to other baselines for CamVid (left) and Cityscapes with small initial dataset size (center: 32) and bigger initial dataset size (right: 320). We plot the mean IoU against the number of image patches acquired in each acquisition step. For visualization reasons we do not show the fully supervised baselines, but report them as 0.73 (0.635) for Camvid with (without) backbone finetuning and 0.618 (0.497) for Cityscapes with (without) backbone finetuning.  (Color best viewed online)} \label{fig:results}
\end{figure*}
We conducted several experiments to evaluate the performance of our method against some active learning strategies from the literature. For comparison, we use \textit{mean Intersection over Union (mIoU)} to measure the performance of the segmentation model, trained on dataset assembled through the active learning strategies. IoU is the ratio of area of overlap between predicted segmentation mask and ground truth label and the area of union between predicted segmentation mask and ground truth label. mIoU is computed by averaging the IoUs of all the classes. We compared our method to following methods from the literature:
\begin{enumerate}
  \item \textbf{Random Sampling (Random):} Na\"ive method of randomly sampling $N$ samples in each acquisition step.
  \item \textbf{Entropy Sampling (Entropy):} An uncertainty sampling strategy capturing the informativeness of samples by computing the entropy \cite{shannon_entropy} as an uncertainty measure.
  \item \textbf{Fully Supervised \cite{deeplabv3, MobileNet}:} Traditional supervised method to train the segmentation model with the full labeled dataset.
\end{enumerate}
We also tried comparing to EquAL \cite{self-consistency}, but found the method to perform poorly in our setting and we did not invest time in its performance tuning. To have a fair comparison with our method, we modify the above methods (except fully supervised) to query and acquire image patches rather than the complete image itself.

\subsection{Datasets} 
We use the publicly available semantic segmentation datasets CamVid \cite{camvid} and Cityscapes \cite{Cordts2016Cityscapes} for our experiments. During the acquisition process, instead of obtaining real annotations from an oracle, we mask the already available labels of the query dataset and reveal them once the AL algorithm selects those patches.

\textit{CamVid} \cite{camvid} is a dataset which is composed of street scene view images of size $360\times480$ with 11 semantic classes. It contains 367 (training), 101 (validation), 234 (test) images with labels. We divide each image into 16 patches of dimension $90\times120$. 

\textit{Cityscapes} \cite{Cordts2016Cityscapes}, a much larger dataset is also composed of real-world street view images of dimension $2048\times1024$ with 19 semantic classes. The training set consists of 2975 images with semantic labels and the validation set contains 500 images. Also here, we divide each image into 16 smaller patches. Hence the dimension of each patch is $128\times256$ since the width and height of images are bigger than CamVid dataset.

\subsection{Implementation Details} \label{sec:implementation_details}
For the segmentation model, we used DeepLabV3 \cite{deeplabv3} with a MobileNetV2 \cite{MobileNet} backbone pre-trained on ImageNet dataset \cite{ILSVRC15} and implemented using PyTorch \cite{pytorch}. The UMAP implementation is based on \cite{mcinnes2018umap-software} and we use the features extracted from the MobileNetV2 \cite{MobileNet} backbone as training inputs. We train the segmentation model using cross-entropy loss and the Adam optimizer \cite{Adam} with the PyTorch default values. After each acquisition step we train the model for 25 epochs and for 32 epochs after the final acquisition step, and chose the best model among them. Note that we retrain the model from scratch after each acquisition step. In all the experiments, the batch size is set to 8, the learning rate is set to 0.001 and no learning rate schedule is used. Moreover we do not use any weight decay and no data augmentation techniques were employed. We repeated each experiment three times with different random choices for the initial labeled dataset $D_{L}$. At the end, we reported the average value of mean IoU score on the validation dataset along with the standard deviation.

The fully supervised baselines were trained with the full training data of CamVid and Cityscapes, respectively. We trained the model for 60 epochs with a batchsize of two; note that we are here using the full images and not just the patches. Moreover we used a learning rate of 0.0008 and used a learning rate decay of 0.1 every 20 epochs. We report the mIoU score on the validation dataset for both setting freezing and finetuning the MobileNetV2 backbone.

\begin{table*}[t]
\centering
 \begin{tabular}{| c | c |c |c |c | c | c | c | c | c | c |} 
 \hline
  & \textbf{finetune} & \textbf{\# params} & \textbf{Init. bs} & \textbf{qs} & \textbf{perc}. & \textbf{Random} & \textbf{Entropy}  & \textbf{MEAL} & \textbf{MEAL-FT} & \textbf{Supervised}\\ 
 \hline
 \textbf{CamVid} & no & 4.0e6 & 32 & 32 & 5\% & 0.509 & 0.516  & 0.558 & \textbf{0.573} & 0.635\\ 
 \hline
 \textbf{CamVid} & yes & 5.8e6 & 32 & 32 & 5\% & 0.52 & 0.528 & 0.54 & \textbf{0.596} & 0.73\\ 
 \hline
 \textbf{Cityscapes} & no & 4.0e6  & 32 & 32 & 1\% & 0.298 & 0.325 & 0.326 & \textbf{0.330} & 0.497\\
 \hline
 \textbf{Cityscapes} & yes & 5.8e6 & 320 & 25 & 1\%& 0.328 & 0.357 & 0.377 & \textbf{0.378} & 0.618\\
 \hline
 \textbf{Cityscapes} & no & 4.0e6 & 320 & 25 & 1\%& 0.277 & 0.310 & 0.312 & \textbf{0.313} & 0.497\\
 \hline
 \textbf{Cityscapes} & yes & 5.8e6  & 32 & 32 & 1\% & 0.333 & 0.384 & \textbf{0.402} & 0.395 & 0.618\\
 \hline
\end{tabular}
\caption{Summary of mean IoU on the validation sets of CamVid and Cityscapes after the final acquisition step. \textit{Perc.} shows the fraction of the acquired dataset compared to the full dataset size, \textit{Init. bs} is the initial dataset size for the seed labeled dataset, \textit{qs} is the query size and \textit{\# params} gives the number of trainable parameters. Best method is highlighted in bold.}
\label{tab:results}
\end{table*}

\subsection{Experimental Results}
Based on some initial hyperparameter searches, we settle on using two initial randomly sampled images and split each into 16 patches. The total amount of 32 labelled patches form our initial dataset. At each acquisition step, we (i) split the input images into 16 patches and (ii) query 32 patches from the unlabeled set based on the active learning strategy at hand. In MEAL (MEAL-FT) this acquisition is based on first extracting the 200 most informative patches, before extracting the final 32 patches using the clustering of UMAP embeddings. We do this in order to guarantee diversity in the acquired samples at the end of each acquisition step. These final queried samples are added to the training set along with the corresponding labels and a new model is trained on this updated training set. We also experiment with a larger initial sample of 20 images and 16 patches each in the case of Cityscapes. In this case, we modify the query sample size to 25 patches. Moreover we run ablations with finetuning and freezing the MobileNetV2 backbone. All results are summarized in Table~\ref{tab:results}.

In Figure~\ref{fig:results} (left), we plot the average mean IoU on the CamVid validation set with respect to the effective training dataset size after new patches have been acquired after each acquisition step. At the end of all acquisition steps, the total number of patches acquired correspond to approximately 5\% of the total training dataset size. We see that finetuning the backbone gives a marginal lift given the increased numbers of parameters that need to be trained. However in both cases, MEAL and MEAL-FT outperform the baselines given by Entropy Sampling and Random Sampling. We refer to Table~\ref{tab:results} for quantitative results.

The results are slightly different on Cityscapes as shown in Figure~\ref{fig:results} (center, right). Due to time constraints we did repeat the experiments over several random seeds, but limited ourselves to one seed to show trendlines. Here the lift obtained by additionally finetuning the MobileNetV2 backbone is significant. This is true for small and large initial batch sizes. In the case of large initial dataset size, we can also select fewer samples per acquisition step and quickly outperform the baseline strategies with fewer initial samples. This indicates that the initial gradients of the active learner are important and strongly influence later acquisition steps. Moreover, we find that MEAL and MEAL-FT do not outperform the Entropy Sampling strategy, except in the case of backbone finetuning and large initial dataset size. We hypothesize that this could be due to the initial hyperparameters, as we transferred the CamVid settings to Cityscapes. This indicates the importance of well chosen hyperparameters in the case of a frozen backbone, but also shows that finetuning the backbone renders the method less sensitive to the choice of hyperparameters. For quantitative results we refer to Table~\ref{tab:results}.

\section{Discussion \& Future Work} \label{sec:conclusion}

In our work, we conceptualized a new active learning query strategy \textit{MEAL}, within an exploration-exploitation framework for image segmentation. The method combines informativeness measures, based on entropy sampling, with representative measures, based on clustering methods. Our method extracts low-dimensional manifold embeddings from feature vectors of unlabeled data points obtained using UMAP which maintains the structural similarity between the data points in lower dimensions. This allows us to efficiently cluster the datapoints using a K-Means++ approach. We conducted experiments on semantic segmentation datasets of different sizes and analyzed the performance of our algorithm. The experiments on CamVid and Cityscapes dataset outperformed existing methods that were adopted to fit to our image segmentation task and validate the correctness of this approach. Moreover, we showed that finetuning the backbone of the segmentation network improved the method but is not uniform over datasets, which is likely due to the usefulness of the representation for the dataset at hand. We note that while MEAL-FT was designed as an intermediate ablation step between MEAL and the Entropy Sampling method we do not see the expected interpolating behavior. This could be due to a too large sample size for selecting the informative subsample of patches. In this case, the large subsample already contains enough representative power of the full dataset and hence we would not see a loss in global context.

There are various directions that can be investigated for further improvement of the approach which we leave for future work. As the current experiments did not perform exhaustive hyperparameter tuning it would be worthwhile to understand the dependence of the algorithm on factors such as the initial batch size, the query size as well as the UMAP embedding dimension. Moreover, the use of UMAP as a dimensionality reduction technique is a sound choice, but certainly not the only one and a more exhaustive comparison to existing active learning approaches for image segmentation should be done. Finally, as seen in the work by \cite{asknlearn}, prediction calibration that provides reliable gradient embeddings needed for querying, could be incorporated in the uncertainty sampling step of our method to obtain calibrated Softmax values.

\section{Author contributions} \label{sec:ac}
D. Sreenivasaiah and T. Wollmann contributed to the design of the method and research. D. Sreenivasaiah implemented the approach.  D. Sreenivasaiah and J. Otterbach performed the analysis. T. Wollmann supervised the work. All authors discussed the results and wrote the manuscript.

\section*{Acknowledgements}
The research leading to these results is funded by the Federal Ministry for Economic Affairs and Energy (BMWi) within the project "AI Safeguarding – Methods and measures for safeguarding AI-based perception functions for automated driving" (\#19A19005Y). The authors would like to thank the consortium for the successful cooperation.

{\small
\bibliographystyle{ieee_fullname}
\bibliography{egbib}
}

\end{document}